\documentclass[a4paper]{article}

\usepackage{INTERSPEECH2022}
\usepackage{caption}
\usepackage{subcaption}
\usepackage{enumitem}

\title{Using Deepfake Technologies for Word Emphasis Detection}

\name{Eran Kaufman$^1$, 
Lee-Ad Gottlieb$^2$}

\address{
   $^1$Shenkar College\\
 $^2$Ariel University}
 \email{erankfmn@gmail.com, leead@ariel.ac.il}

\begin{document}

\maketitle
\begin{abstract}
 In this work, we consider the task of automated emphasis detection for spoken language. This problem is challenging in that emphasis is affected by the particularities of speech of the subject, for example the subject's accent, dialect or voice.

To address this task, we propose to utilize deep fake technology to produce an emphasis-devoid speech for this speaker. This requires extracting the text of the spoken voice, and then using a voice sample from the same speaker to produce emphasis-devoid speech for this task. By comparing the generated speech with the spoken voice, we are able to isolate patterns of emphasis which are relatively easy to detect. 
\end{abstract}


\section{Introduction}\label{sec1}                                                              
We as humans have developed a deep sensitivity to the `music' of speech, meaning its stress, rhythm and intonation. Intonation in particular may be used to express wonder, cynicism or emphasis, and any one of these may alter (or even completely reverse) the meaning of a sentence.

Let us take for example the simple sentence `I did not take your bag.' Placing emphasis on different words of the sentence can affect its overall meaning: Emphasizing the subject of the sentence -- `\emph{I} did not take your bag' -- implies that the bag may still have been taken, but by someone else. Emphasis on the possessive adjective -- `I did not take \emph{your} bag' -- implies that I did take a bag, only a different one. And emphasis on the object -- `I did not take your \emph{bag}' -- implies that I took a different object of yours.

Establishing the correct emphasis in a spoken sentence is therefore central to a correct interpretation of that sentence. Indeed, written language has long ago adopted tools to convey emphasize or meaning, such as italicization, punctuation marks, and the more recent use of emoji symbols. Hence, understanding and classifying word emphasis is a potentially important task for fields related to human-machine interaction, for example machine translation, spoken information retrieval, automated question answering, sentiment analysis and speech synthetics.

\smallskip

\noindent \textbf{Our contribution.}
The task of automated emphasis detection is complicated due to the fact that different languages, dialects or accents already feature inherent differences in emphasis.
In addition, different voices resonate at different frequencies. 
Hence, this makes our task speaker specific.
We propose to address this problem by employing deepfake techniques: Given a spoken statement upon which emphasis must be determined, we utilize deepfake methods (along with a speech sample from the same speaker) to automatically generate an 'emotionless' version of this query statement, that is a version which mimics the speaker reciting the query statement with no particular emphasis. Then by comparing the original query statement and its synthesized emphasis-devoid version, we can identify emphasized parts of speech in the query statement.

An overview of our computational approach is as follows:
Our detector is built by composing several separate modules.
A voice encoder processes the speech sample to produce a representative data vector capturing the speaker's voice characteristics.
Given the query statement, a speech-to-text (STT) module generates text from the spoken sentence. 
Then a text-to-speech (TTS) module uses the embedded data vector and the text of the spoken sentence to generate an audio waveform of the same text as if produced by the same speaker, but devoid of any special emphasis. This constitutes the deepfake synthesized version of the speech. 
Finally, an analyzer will compare the query statement and its deepfake. As these two differ solely in their emphasis, this final step finds the emphasized words.

We remark that our work is in harmony with the theme of inclusivity of INTERSPEECH 2023, and has the potential to facilitate interaction between peoples of different dialect and idiosyncrasies of speech.

\begin{figure*}
  \centering
  \includegraphics[width=\linewidth]{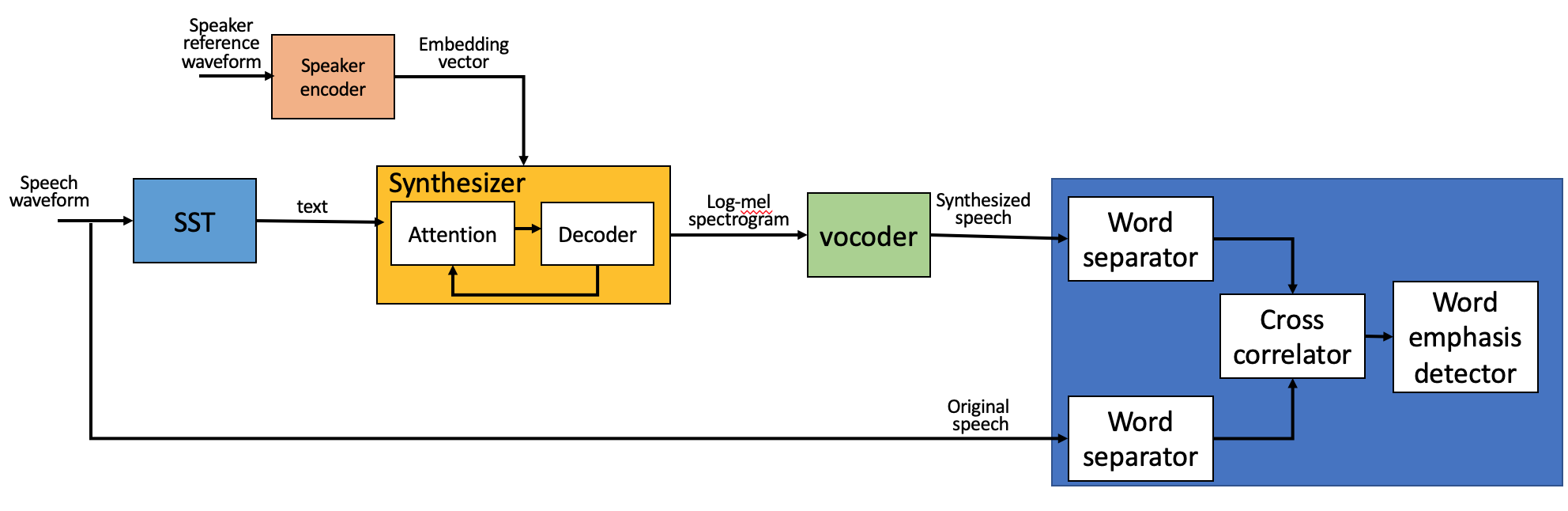}
  \caption{Algorithm work flow.}
  \label{fig:work}
\end{figure*}

\section{Related Work}\label{sec:rel}
Prosody and word emphasis are the subjects of significant research in the field of speech correction, in particular as relates to speech of non-native speakers. It has also received much attention in the field of neural TTS synthesis, where attention to emphasis can yield more expressive speech.

Intonation models, such as the Fujisaki \cite{Fujisaki1983}, 
Hirst \cite{hirst1992prediction}, 
Rise/Fall/Connection (RFC) \cite{taylor1994rise}
and Tilt models \cite{taylor1998tilt},
aim to provide linguistically meaningful interpretations to an utterance.
The Fujisaki model works on the F$_0$ (fundamental frequencies) contour by applying a pair of filters to generate the phrase and accent components of speech, and then adding these to a base frequency value.
By specifying different amplitudes and durations, the model is able to  successfully detect different types of accents.
In the Hirst model, the F$_0$ contour is first encoded by a number of target points using a fitting algorithm.
It is then classified into different phonological descriptions.
Similar to the Hirst model, the RFC model attempts to split the F$_0$ contour into three different categories, a rise or fall in intonation, or a neutral connection.
In the Tilt model, amplitude, duration and tilt are used for describing the intonation shapes of rise or fall, or a rise followed by a fall. 

Basic components of an intonational events include pitch accents and edge tones \cite{ladd2008intonational}. Pitch accents are associated with syllables and signify emphasis, while edge tones occur at the edges of the phrase and give cues such as continuation, question or statement. 
Kun et al.\ \cite{kun} used intonation detection in order to detect errors in English speech, and to then provide corrective feedback to speakers of English as a second language. They developed a pitch accent detector based on a Gaussian mixture model, and used features based on energy, pitch contour and the vowel duration.

There have been several relevant contributions in the field of neural TTS, with the overarching goal of improving generated prosody.
Several variational \cite{HsuZWZWWCJCSNP19,8683623} and non-variational \cite{Skerry-RyanBXWS18,WangSZRBSXJRS18}
models have been suggested for learning latent prosodic representations.
One line of work proposes methods for low-level prosody control \cite{vioni2021prosodic,GongWLG0D21}, while another exploits various syntactic and semantic features to generate context-suited prosody
 \cite{GuoSHX19a,SongLZ0M21,HayashiWTTTL19,9054337}.
 Bai et al.\ \cite{BaiKZ22} used a separate model to extract semantic features such as questions from text alone, and then fed these new features into a chosen vocoder.
Similarly, Mass et al.\ \cite{MassSMHSLK18} suggested incorporating a word emphasis predictor based on text alone. Their predicator is based on recurrent neural networks (RNNs), and its output was fed to a TTS module.
They found that word emphasis patterns are both speaker specific and difficult to identify using only the speaker's voice. This motivated their use of an independent emphasis predictor.
We aim to solve this problem by incorporating a sample of the speaker's own voice into the learning process, and use this to generate emphasis-devoid baseline waveform for emphasis comparison.

 \begin{figure*}
  \centering
  \includegraphics[width=\linewidth]{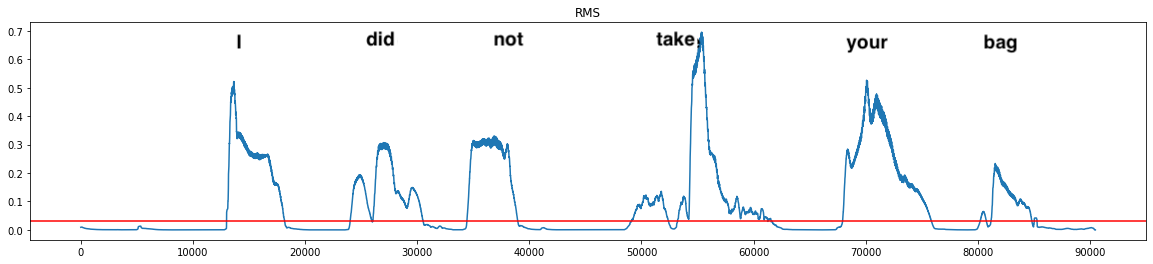}
  \caption{RMS sliding window and word separation.}
  \label{fig:rms}
\end{figure*}
 
\section{Our Work}\label{sec:our}
 We present a new deepfake-based approach for emphasis detection. Our algorithm is given a speech sample from a target speaker, and uses this to familiarize itself with the ambient properties of this speaker. Then given a spoken query statement, the algorithm extracts the text of the query, and produces a `vanilla' TTS version of this text (that is, TTS with no specific emphasis). This emphasis-void speech is then compared to the query statement.

\smallskip
 
\noindent \textbf{Background.} Recent vanilla neural TTS synthesis technologies have achieved realistic synthetic speech generated
from a very small sample of a speaker's voice \cite{RenRTQZZL19,KimKS21,jia2018transfer}.
These TTS models are based on deep neural networks, and are trained 
 using an encoder-decoder architecture. They map input characters or phonemes to acoustic features (for example, mel-spectrograms) or directly to the waveform. The acoustic features can be converted into waveforms via vocoders \cite{OordDZSVGKSK16,YangYLF0X21}.
 
Our work is based primarily on the SV2TTS TTS architecture \cite{jia2018transfer}. This specific architecture is composed of three independently trained neural networks:
\begin{itemize}[leftmargin=*]
\item A speaker encoder (based on \cite{wan2018generalized}), which uses a sample of the speaker's voice to compute a fixed size embedding vector.
\item A sequence-to-sequence synthesizer (based on \cite{shen2018natural}), which constructs a mel-spectrogram from
a sequence of grapheme or phoneme inputs, conditioned on the  embedding vector.
\item An autoregressive WaveNet vocoder \cite{oord2016wavenet}, which converts the mel spectrogram into time-domain waveforms.
\end{itemize}

\noindent \textbf{Our construction.}
We will utilize all three of these neural networks. The encoder is used to produce an embedding vector representing properties of the speaker's voice, that is a `voice print.' This will later be used to produce a deepfake of the speaker reciting the query statement.
The synthesizer is fed by text sequences concatenated with the speaker's embedding  vector to create the log-mel spectrogram.
The log-mel spectrogram is fed into the vocoder to output a synthetic waveform. To these we will add a detector to compare the synthetic and the original waveforms and identify pitch or skew accent.
Our word emphasis detector is composed of five distinct ordered parts:

\smallskip
\noindent \textbf{Step 1: Encoder.}
The above encoder utilizes a voice sample provided by the speaker to create an embedding vector representing the voice properties of the speaker.

\smallskip
\noindent \textbf{Step 2: Speech to text.}
The speaker's query statement is inputted into an STT module, which extracts the text of this statement.

\smallskip
\noindent \textbf{Step 3: Text to speech.}
The TTS module uses the synthesizer described above.
Both the text produced from the STT step and the embedding vector produced by the encoding step are fed to the synthesizer, which then produces a waveform. 

This waveform is an emphasis-devoid deepfake of the speaker reciting the query statement.
We recall that vanilla neural TTS systems are not capable of synthesizing emphasis due to the loss of sentiment information \cite{bai22c_interspeech}. This computed waveform serves us as the baseline for the task of emphasis detection.

\smallskip
\noindent \textbf{Step 4: Waveform comparison.}
Having computed the synthesized speech, we can compare it to the spoken query statement, to determine which word or words are emphasized. Our comparison technique is detailed in Section \ref{sec:compare} below.

\subsection{Comparison between waveforms} \label{sec:compare}
Our premise is that the synthesizer can produce a reasonable imitation of emphasis-devoid speech of the speaker. The emphasis of a word by the speaker may differ from the synthesizer waveform in that the speaker's word is pitched or skewed relative to the normal voice produced by the synthesizer. Hence, a cross correlation test between the respective spectrograms of these two waveforms may allow us to identify the special emphasis made by the speaker.

To effectively compare the two waveforms, we need to first separate both the synthetic and query speech into their distinct words. This is done using a sliding root mean square (RMS) window, while applying a low threshold to distinguish between spoken and silent parts of the speech (see Figure \ref{fig:rms})\cite{QiH93}. We then compute the fast Fourier transform (FFT) for each individual word, and compare for each word its two respective spectrograms corresponding to the synthesized and spoken speech.
We focus on the two distinct modes of emphasis mentioned above: 

\begin{itemize}[leftmargin=*]
\item 
The first is \emph{pitch}, meaning that the speaker's emphasis of a word is accomplished by 
modulating regular speech into a higher (or sometimes lower) tone. 
 In this case, the general shape of the spectrogram remains the same, but its central frequency shifts. This is identifiable by the peak of the cross correlation of the two spectrograms. 
\item
The second is \emph{skew}, wherein the speaker modulates the voice up and down to emphasize a word. Here the spectral distribution is significantly different from the auto-generated waveform, and its total energy is spread over a wider range of the spectrum. In this case the cross correlation between the two spectrograms is low for all frequency shifts.
\end{itemize}

\begin{figure}
\centering
\begin{subfigure}{\linewidth}
  \centering
  \includegraphics[width=\linewidth]{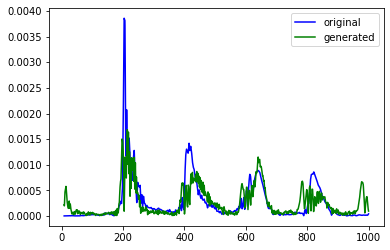}
  \caption{FFTs of the original and generated wave forms}
  \label{fig:bag1}
\end{subfigure}%
\newline
\begin{subfigure}{\linewidth}
  \centering
  \includegraphics[width=\linewidth]{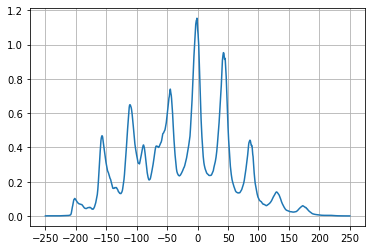}
  \caption{The correlation between the two FFTs}
  \label{fig:bag2}
\end{subfigure}
\caption{Comparison of the FFTs for the word `bag.'}
\label{fig:bag}
\end{figure}

\begin{figure}
\centering
\begin{subfigure}{0.5\linewidth}
  \centering
  \includegraphics[width=0.9\linewidth]{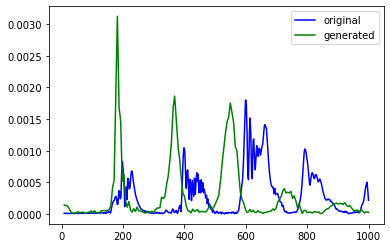}
\end{subfigure}%
\begin{subfigure}{0.5\linewidth}
  \centering
  \includegraphics[width=0.9\linewidth]{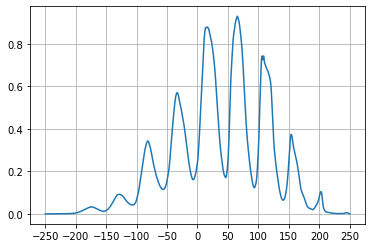}
\end{subfigure}
\caption{Comparison of the FFTs for the word `take.'}
\label{fig:take}
\end{figure}

Detection of differences due to pitch is illustrated by the comparison of Figures \ref{fig:bag} and \ref{fig:take}:

Figure \ref{fig:bag} illustrates the above comparison for the word `bag' in the sentence `I did not \emph{take} your bag' (i.e., where the word `take' and not `bag' is emphasized.)
The comparison is between the spoken and generated waveforms' FFT.
One can see that the spectral analysis of the two waveforms are quite similar, and this is due to the fact that the word `bag' was not emphasized in this query.
The figure showing the cross-correlation between the two FFTs shows that the peak is close to zero, implying a relatively high correlation between the two waveforms.

Figure \ref{fig:take} illustrates the comparison of the word `take' in the same sentence (where 'take' was indeed emphasised). It is readily seen that the synthesized and spoken spectrum of the waveforms differ significantly. The corresponding cross correlation demonstrates a shift of the peak correlation of the signal by about $80Hz$.

The general shape of these two spectrograms does not differ significantly, and so by applying a threshold on the frequency shift of the cross correlation it is possible to identify if the word was emphasized. The correlation and FFTs are normalized by the wave's total energy.

\begin{figure}
  \centering
  \includegraphics[width=0.9\linewidth]{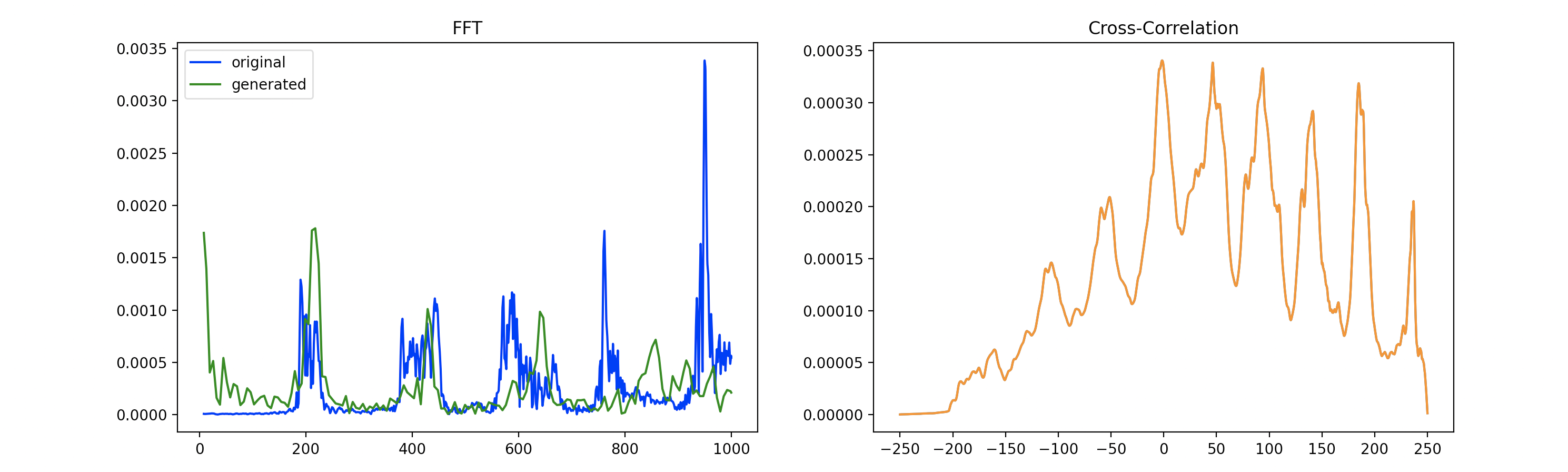}
\caption{Comparison of skew words: (a) FFTs of the original and generated wave forms. (b) Their correlation.}
\label{fig:ill}
\end{figure}

For skew we applied a low threshold on the cross correlation peak amplitude.
When there is no correlation at all this is also a signal of word emphasis (see Figure \ref{fig:ill}).

Figure \ref{fig:work} illustrates the workflow of the algorithm.
The recorded waveform is inputted into the STT module. The extracted text and the embedded vector of the speaker are fed into the synthesizer, which then creates the synthetic mel-spectrogram which turns into a waveform by the vocoder. Both the original and synthesized wave forms are fed into a decomposition module which separates the speech into its individual words. Corresponding words are compared using the FFT cross correlation module, and this determined whether or not the word was emphasized.


  
\section{Implementation and experiments}\label{sec:exp}
As already mentioned in Section \ref{sec:our} above, our encoder and decoder are adapted from the SV2TTS architecture \cite{jia2018transfer}. 
which was in turn, based on the recurrent sequence-to-sequence Tacotron2 network \cite{ShenPWSJYCZWRSA18}, extended with an attention network to support multiple speakers, similar to the scheme suggested for Deep Voice2 \cite{GibianskyADMPPR17}. 

We used the sample-by-sample autoregressive WaveNet \cite{oord2016wavenet} as a vocoder to invert synthesized mel-spectrograms emitted by the synthesis network into time-domain waveforms. This architecture is composed of $30$ dilated convolution layers, similar to what was described in \cite{shen2018natural}. The network is not directly conditioned on the output of the speaker encoder. The mel-spectrogram predicted by the synthesizer network captures the information needed to produce a multi-speaker vocoder. To train the speech synthesis and vocoder neural networks, we used the VCTK \cite{VCTK} dataset, which contains 44 hours of speech from 109 speakers. We downsampled the audio files to 16 kHz, and trimmed leading and trailing silence sequences.

Our word emphasis predictor, described in Section \ref{sec:our} above, computes the word by word cross-correlation between the generated and the original FFTs. Since the number of samples for the generated and original words are not of the same in length, a simple linear interpolator is applied in the frequency domain. 

For our experiments, we constructed a dataset of $100$ different voice samples: Five different speakers recited five different sentences, each sentence with word emphasis on one of four different words. The five sentences are:
\begin{enumerate}
    \item 
    ``\underline{I} did not \underline{take} \underline{your} \underline{bag}.'' 
    \item
    ``\underline {Hello}, \underline {this} is \underline {our} \underline{intonation}  project.'' 
    \item
''There are very \underline{few} \underline {black} rhinos \underline{left} in \underline {Africa}.'' 
    \item
''\underline{I} \underline{saw} her \underline{face} under the \underline{hood}.'' 
    \item
    ''Why did \underline{you} give \underline{Sarah} the \underline {sandwich} with \underline{mustard}.'' 
\end{enumerate}
The above underlined words were the ones given emphasis.
We obtained an accuracy, precision, recall, and F$1$ score of $92$, $89.14$, $89.33$, and $89.23$, respectively.

\begin{figure}
  \centering
  \includegraphics[width=\linewidth]{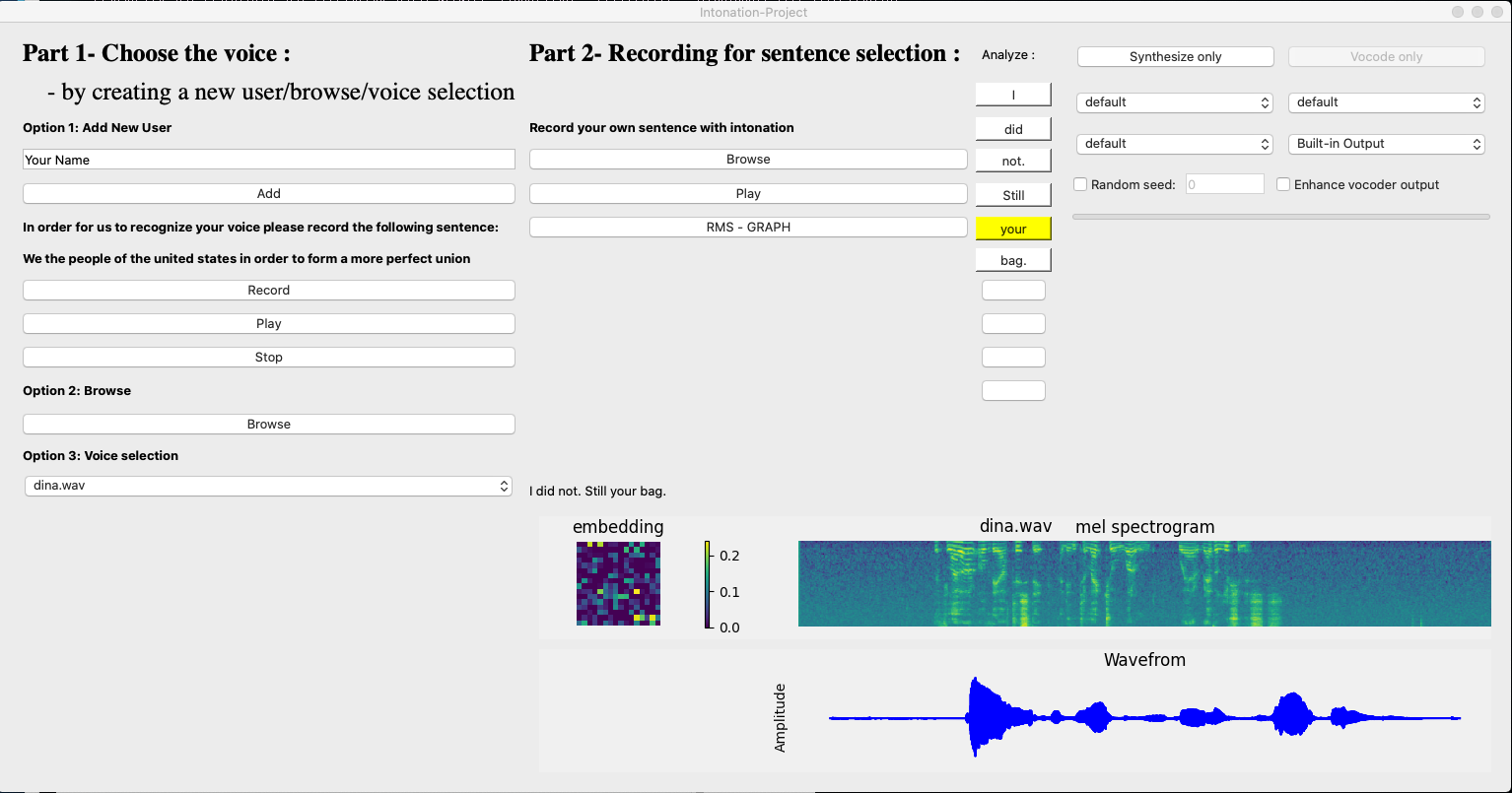}
  \caption{RMS sliding window and word separation.}
  \label{fig:panel}
\end{figure}

The project open source code can be found online.\footnote{https://anonymous.4open.science/r/Intonation-Project-215B}
It runs as a python application with 3 distinct parts:
(i) Configuration of a user using live or recorded voice.
(ii) Recording a sentence from that same user to create a synthetic voice.
(iii) Word emphasis - the recording is converted into text and a separation is applied. Each word is placed in a different box, with emphasized word boxes highlighted.
Pressing a box will open the spectrum analysis of the original and synthesized words, alongside the cross correlation between them (See figure \ref{fig:bag}).
The control panel is demonstrated in figure \ref{fig:panel}.
The original speech in the time domain is represented in blue,
the mel-spectrogram of the synthesized speech as outputted from the decoder is found above, and the embedding vector of this specific user 
is found alongside.
The output result highlights the word `take' which was emphasized in this specific example.
Videos demonstrating the use of the application can be found online.\footnote{%
https://www.youtube.com/@intonationdetection-kl7np
}

\section{Conclusions and Future Work}
In this paper, we presented the layout and empirical results for our word emphasis detector.
As we have described above, this problem is especially challenging in that emphasis is affected by dialect and accent, and also different voices may differ significantly in their resonance. For this problem we developed a novel approach using deep fake technology to produce an emphasis-devoid speech for this speaker. We used  a double conversion from speech to text and back to speech again. By comparing the generated and spoken voice, we are able to isolate patterns of emphasis which are relatively easy to detect.

For future work, we intend to use our technique not only to detect emphasis, but also to cluster and classify different emotions for the purpose of sentiment analysis.

\newpage
\newpage

\bibliographystyle{IEEEtran}

\bibliography{intonation}

\end{document}